\documentclass{article}

\usepackage{arxiv}

\usepackage[utf8]{inputenc} 
\usepackage[T1]{fontenc}    
\usepackage{hyperref}       
\usepackage{url}            
\usepackage{booktabs}       
\usepackage{amsfonts}       
\usepackage{nicefrac}       
\usepackage{microtype}      
\usepackage{lipsum}
\usepackage{graphicx}
\graphicspath{ {./images/} }
\usepackage{xcolor}
\usepackage{color, colortbl}
\definecolor{primo}{rgb}{0,0.9,0.1}
\definecolor{secondo}{rgb}{0.7,1,0.7}
\definecolor{terzo}{rgb}{0.9,1,1}
\def\myalgoname{TUCaN }
\def\primarycapsdown{\texttt{PCD}}
\def\primarycapsup{\texttt{PCU}}
\def\doubleblockdown{\texttt{DBD}}
\def\doubleblockup{\texttt{DBU}}

\def\mybn{\texttt{BN}}
\def\myrelu{\texttt{ReLU}}

\def\myconv{\texttt{Conv}}
\def\mymaxpool{\texttt{MaxPool}}
\def\myTconv{\texttt{TransposeConv}}

\def\myflatten{\texttt{Flatten}}
\def\myupsample{\texttt{UpSample}}

\title{TUCaN: Progressively Teaching Colourisation to Capsules}

\author{
 Rita Pucci \\
  Machine Learning and Perception Lab\\
  University of Udine, Udine\\
  ORCID:0000-0002-2970-1180 \\
  \texttt{rita.pucci85@gmail.com} \\
   \And
Niki Martinel \\
  Machine Learning and Perception Lab\\
  University of Udine, Udine\\
  ORCID: 0000-0002-6962-8643 \\
  \texttt{niki.martinel@uniud.it} \\
}

\begin{document}
\maketitle
\begin{abstract}
Automatic image colourisation is the computer vision research path that studies how to colourise
greyscale images (for restoration). Deep learning techniques improved image colourisation yielding
astonishing results. These differ by various factors, such as structural differences, input types,
user assistance, etc. Most of them, base the architectural structure on convolutional layers with no
emphasis on layers specialised in object features extraction.
We introduce a novel downsampling-upsampling architecture –named TUCaN (Tiny UCapsNet)–
that exploits the collaboration of convolutional layers and capsule layers to obtain a neat colourisation
of entities present in every single image. This is obtained by enforcing collaboration among such
layers by skip and residual connections.
We pose the problem as a per-pixel colour classification task that identifies colours as a bin in a quantized
space. To train the network, in contrast with the standard end-to-end learning method, we propose
the progressive learning scheme to extract the context of objects by only manipulating the learning
process without changing the model. In this scheme, the upsampling starts from the reconstruction
of low-resolution images and progressively grows to high-resolution images throughout the training
phase. Experimental results on three benchmark datasets show that our approach with ImageNet10k
dataset outperforms existing methods on standard quality metrics and achieves state-of-the-art
performances on image colourisation. We performed a user study to quantify the perceptual realism
of the colourisation results demonstrating: that progressive learning let the TUCaN achieve better
colours than the end-to-end scheme; and pointing out the limitations of the existing evaluation metrics.
\end{abstract}


\section{Introduction}
\begin{figure}[h]
  \centering
  \includegraphics[width=0.5\linewidth]{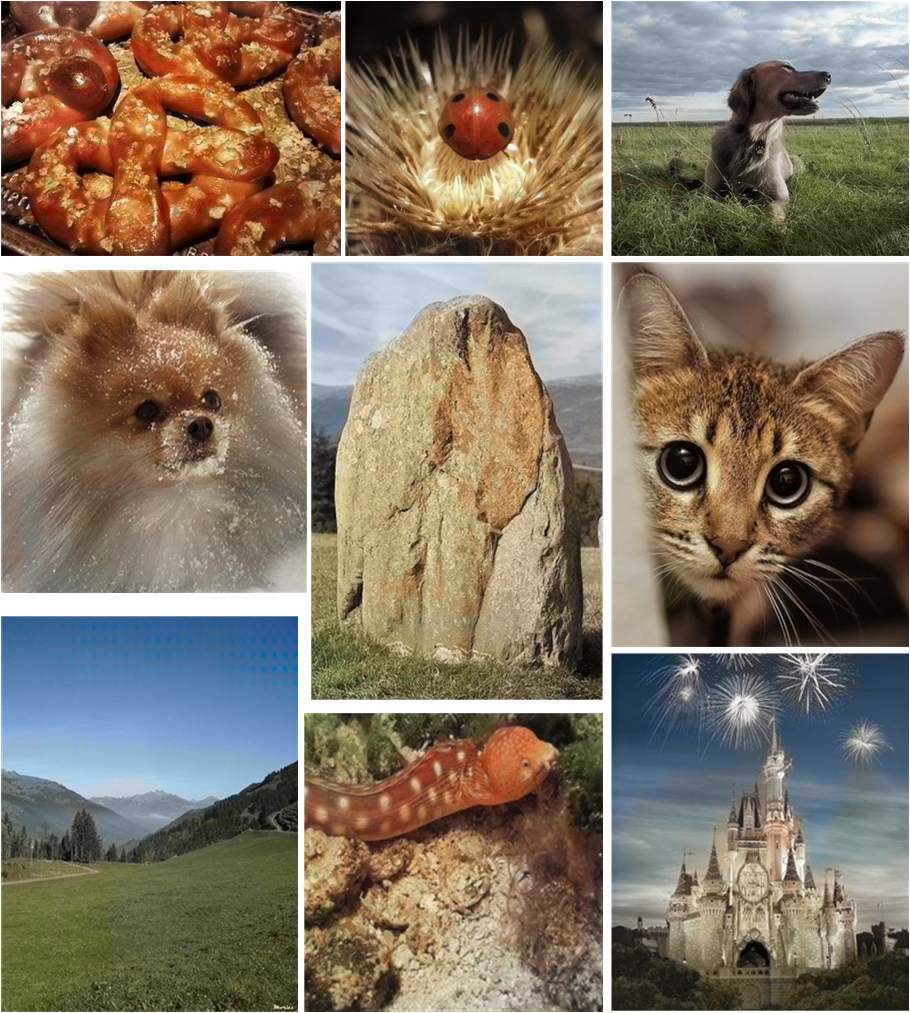}
  \caption{\myalgoname trained in progressive learning, is capable of producing natural and colourful results within different context. Here we showcase samples of animals, landscapes, food images coloured with our proposed approach. }
  \label{fig:Samples}
\end{figure}
In this paper, we refer to \textit{colourisation problem} as the willing of predict natural colourisation of a greyscale image. The challenge of this problem is due to the variety of image conditions that need to be dealt via a single algorithm. In an automatic colouring process, the input of the algorithm is an image where the chrominance channels are lost, and only the luminance channel provide information for colour reconstruction. The absence of information about the colourisation makes the colourisation problem severely ill-posed and underlines the multi modality of colourisation. Although the semantics of the scene may be helpful in reducing the plausible colourisation of elements, for example, grass is usually green, clouds are usually white, and the sky is blue. We do not find these  semantic priors for objects such as t-shirts, desks, and many other observable items. Moreover, the colourisation problem also pertinent to changes in illumination, variations in viewpoints, and occlusions that make hard the semantic comprehension of objects from the algorithm.
The rapid development of deep neural networks opens to new proposed models for image colourisation. These models ranging from the early straight forward networks (e.g.,~\cite{cheng2015deep}) to the more recent carefully designed works (e.g.~\cite{Su2020CVPR}) raising the state of the art (SOTA) bar. In particular, we take inspiration by some of the works that deal with the colourisation problem deep neural networks in the last years such as: ~\cite{cheng2015deep,iizuka2016let,zhang2016colorful,larsson2016learning,deshpande2017learning,guadarrama2017pixcolor,isola2017image,he2018deep,ozbulak2019image,mouzon2019joint,Su2020CVPR}. These proposed models considered image-level or entity-level features, or both at the same time but independently, thus neglecting the importance of the interaction between the global content (image-level features) and the entity instances (entity-level features) in an image. 

We identify as image-level features the features that provide the general sense of the image, and as entity-level features the detailed features that focus on the entities in the image. Following this idea, \textit{we introduce a novel approach that performs automatic colourisation by combining the features from the image context with the object instances, \myalgoname}. 
This model consists of a collaboration among of convolutional and capsules layers~\cite{sabour2017dynamic}.The convolutional layers identify and extract features which carry information about the image context, and capsules layers, by the means of the routing by agreement routine, capture the presence and the features of entities. The collaboration is implemented by residual and skip connections over aligned layers, in U shape architecture. 

The model presented is trained to deal with learning a distribution of colours from two prospective: (i) the colours are quantised and identified over a fixed amount of classes, and the model predicts the colourisation pixelwise; (ii) from the predicted quantised colourisation, the model predicts the two missing chroma channels.

At SOTA, all of the works proposed in colourisation have an end-to-end learning scheme with the same input and output sizes. However, this learning scheme causes some issues in training the upsampling for the colour prediction. In particular, the training procedure concern too many parameters to be learnt at ones and that combined with the multi modality of the colourisation problem cause the prediction of desaturated colours, this behaviour is observable in a final brownish colourisation of the image. In this paper, we apply a modified, enhanced version of the progressive learning~\cite{karras2017progressive}, that grows  progressively while epochs go on. We compare \myalgoname trained in progressive learning with \myalgoname trained in end-to-end learning and we discuss the results from a quantitative and qualitative point of view. To the best of our knowledge this is the first time that progressive learning is used for colourisation purposes. 

We have conducted a comprehensive evaluation of our model on three large scale benchmarks datasets, namely ImageNet10k~\cite{russakovsky2015imagenet} shown in some samples in Fig.~\ref{fig:Samples}, COCOStuff \cite{caesar2018coco} and Places205~\cite{zhou2014learning}.
Results demonstrate that, on ImageNet10k, our approach outperforms existing works in terms of the image colourisation quality using common image-quality metrics, while the method appear competitive for the other two datasets. We prove that the proposed method performs better that the closest colourisation method at SOTA based on capsules.
We also evaluate the qualitative performance of the proposed progressive learning approach compared to the end-to-end approach through a large scale human-based evaluation study. Finally we tag some interesting points of discussion: Are the metrics proposed for colourisation the things that tip the scale of plausibility in a colourisation model?

\begin{figure*}[!h]
  \centering
  \includegraphics[width=\textwidth]{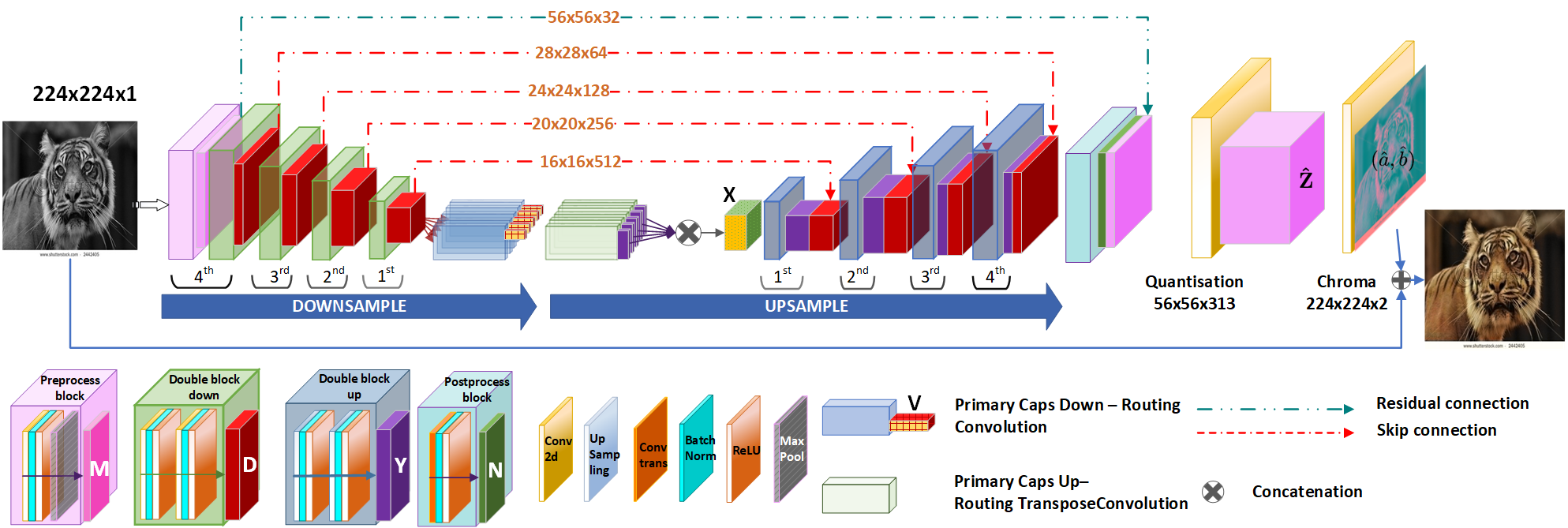}
  \caption{Structure in details of the upsampling and downsampling phases of \myalgoname and the loss computation procedure for last layer of colourisation.}
  \label{fig:TUCaN}
\end{figure*}

\section{Related works}
The aim of colourisation is to convert a grayscale image to a colour image, typically captured in previous decades, when technological advancements were limited.
Image colourisation has been theatre of significant research efforts over the past few years where we have seen a huge shift in interest towards deep learning-based models due to their success in a number of different application domains. Generative models~\cite{isola2017image, nazeri2018image,vitoria2020chromagan} were exploited together with pre-trained classifiers to obtain the information needed for predicting a plausible colourisation. Other approaches~\cite{zhang2016colorful,larsson2016learning,zhao2018pixel,zhang2017real} focused on multi-task models and single pixel significance to pic up a right colourisation for the image but without taking into consideration the importance of entity level features. Works such as  ~\cite{iizuka2016let,zhao2018pixel} brought up the use of semantic labels, idea that exploited to capture entity-level features and, together with image-level features, improve the colourisation. The importance of semantic information and entity level features is committed also in~\cite{zhang2016colorful}, where semantic interpretability is obtained by a cross-channel encoding scheme, later exploited in~\cite{mouzon2019joint} with a pre-trained classification model. Recently, \cite{ozbulak2019image} proposes to embrace the capsules concept to catch the idea of entities features, while \cite{Su2020CVPR} support the main convolutional model with a detection model that emphasises the entity details.

At the SOTA, there are not previous works that involve the use of progressive learning for colourisation of images. Works like \cite{karras2017progressive, natsumi2019} propose progressive learning applied for generation of images of humans in fashion, the models provide a progressively high resolution image that is evaluated by discriminators. From this idea of progressive learning, \cite{kang2020pl} proposes a model trained with progressive learning in multi tasking for salient object detection methods. At each step of progressive learning, the loss function is computed between the predicted salient map and the expected one at the same resolution. 

The works in~\cite{ozbulak2019image} and~\cite{Su2020CVPR} are the closest to our approach, we demonstrate that the \myalgoname outperforms qualitatively and quantitatively both works.

We present the following contribution:
(A) we introduce a single architecture that
(i) captures entity-level features through capsules without using any prior or external knowledge to detect and extract object information; 
(ii) obtains the spatial information discarded by capsules through image-level features from convolutional layer;
(iii) has a strong collaboration among layers by the means of skip  and residual connections.
(B) we propose the progressive learning procedure for the colourisation task and we demonstrate that by progressive learning we obtain high quantitative results in terms of metrics and qualitatively vivid and sharp colourisation.
(C) we apply a combination of a colour quantisation loss~\cite{zhang2016colorful} to learn a distribution of colours and a colour error loss that allows us to deal with multiple natural colourisations of a same/similar object. 

To the best of our knowledge, the proposed architecture, the related idea of convolutional and capsules layers collaboration are new and this is the first application of progressive learning for colourisation. 

\section{Proposed approach}
We propose \myalgoname (Fig.~\ref{fig:TUCaN}), a model trained to predict colours for a grayscale input image. Concretely, starting from the CIELab lightness channel $\mathbf{L}\in \mathbb{R}^{H\times W\times 1}$, \myalgoname learns to predict the corresponding colour channels $(a,b)\in\mathbb{R}^{H\times W\times 2}$. The $(a,b)$ channels represent the chrominance channels used in the training procedure.

\subsection{\myalgoname architecture overview}
\label{cap: arch}
\myalgoname consists of two phases the downsampling and the upsampling (Fig.~\ref{fig:TUCaN}), each of which is made up of convolutional layers blocks and a capsule layers block. The  downsampling extracts features from $\mathbf{L}$, while the upsampling reconstructs colours and resolution of the image. The two phases collaborate because of skip and residual connections.

\textbf{\textit{The downsample phase}} is responsible for learning image-level and entity-level representations.
Learning image-level representation is entrusted to a preprocessing block followed by a set of consecutive \textit{double block down} (\doubleblockdown) layers.
The preprocessing is composed of a \myconv  \mybn  \myrelu  \mymaxpool sequence that perform the first feature extraction (Fig.~\ref{fig:TUCaN}) and it achieve a reduction of resolution by a factor of two, output $\mathbf{M}$. In the consecutive \doubleblockdown{ }, we followed a common practice~\cite{hadji2018we} by which image-level features can be obtained through a hierarchical structure of repeated convolutional, normalisation, and non-linearity layers. Thus, we let each ~\doubleblockdown{ } be composed of two consecutive sequences of \myconv  \mybn  \myrelu layers. \myalgoname consists in a total of four~\doubleblockdown{ } stacked where $\mathbf{D}^\alpha$ is the output matrix of $\doubleblockdown_\alpha$, for $\alpha=4,\cdots,1$. 

The capsule layer block, named \textbf{\textit{primary capsule down}} (\primarycapsdown), extends the downsampling phase moving from image-level feature extraction into entity-level ones through capsules. The \primarycapsdown{ } is composed of two capsule layers. The first layer of capsules computes $\mathbf{U} = [ \myflatten(\myconv_1(\mathbf{D}^4))^T,$ $ \cdots, \myflatten(\myconv_k(\mathbf{D}^4))^T]$, with each column of $\mathbf{U}$ being the capsule output $\mathbf{u}_i \in \mathbb{R}^k$.
A weight matrix $\mathbf{W}_{ij} \in \mathbb{R}^{k\times \hat{k}}$ is introduced to obtain the second layer of capsules $\hat{\mathbf{u}}_{j|i}=\textbf{W}_{ij}\mathbf{u}_i$, later grouped through the ''routing by agreement'' mechanism \cite{sabour2017dynamic}. We perform three iterations and at each iteration, $\hat{\mathbf{u}}_{j|i}$ are grouped in agreement with the coupling coefficient $\mathbf{c}_i$ to identify clusters of features,i.e., the entity-level feature vector $\mathbf{v}_j \in \mathbb{R}^{\hat{k}}$. Matrix $\mathbf{V}={\mathbf{v}_0,\cdots, \mathbf{v}_j}$ carries information about how strong the capsules agree on the presence of an entity.

\textbf{\textit{The upsample phase}} leverages the entity-level and image-level representations to predict the image colourisation. The \textbf{\textit{primary capsules up}} (\primarycapsup) block decodes $\mathbf{V}$, obtained from \primarycapsdown{ }, to obtain the features about extracted for entities. Concretely, the \primarycapsdown{ } outputs $\mathbf{v}_j$'s contain entity-level features but not the details about their spatial displacement. Since this information is fundamental for the colourisation task, we introduce a mechanism that inverts the \primarycapsdown{ } procedure to reconstruct the spatial information. We introduce a weight matrix $\mathbf{W}^{r}_{ji} \in \mathbb{R}^{\hat{k}\times k}$ connecting each \primarycapsdown{ } output to the \primarycapsup{ } capsules. This computes $\mathbf{u}^{r}_{i} = \mathbf{W}^r_{ji}\mathbf{v}_j$, that are then stacked to obtain $\mathbf{U}^r$ having the same size of $\mathbf{U}$. The resulting $k$ rows of $\mathbf{U}^r$ are then reshaped, processed by $k$ independent \myTconv operators, then concatenated to obtain an output matrix $\mathbf{X}$. 
Thus, $\mathbf{X}$ contains spatial information generated through the entity-level features obtained through capsules. This is composed of four \textit{double block up} (\doubleblockup) designed following the same considerations adopted for the \doubleblockdown, hence composed of two consecutive sequences of \myupsample-\mybn-\myrelu layers.
As shown in Fig.~\ref{fig:TUCaN}, $\doubleblockup^(\alpha+1)$, with $\alpha=1,\cdots,3$ receives as input the concatenation of $\mathbf{D}^\alpha$, from skip connection, and the output of the preceding $\doubleblockup_{\alpha}$, denoted as $\mathbf{Y}^{\alpha}$. Exception for, $\doubleblockup^1$ that processes the de-routed entity-features in $\mathbf{X}$.
The up-sampling operations in each \doubleblockup allow us to reconstruct the $(a,b)$ channels having the same size of the input image. The skip connections enforce exploitation of higher resolution features that can be missing due to the sparsity of the up-sampling operations. Finally the concatenation between the matrices $\mathbf{D}^4$ and $\mathbf{Y}^4$ pass through the postprocessing block that is mirroring the preprocessing block in downsampling phase. The $ \mathbf{N}$ output of postprocessing block is added by residual connection with $\mathbf{M}$. The model learns a colour distribution over pixels (Quantisation layer) that is later exploited to predict the colour channels (Chroma layer). 
 The Quantisation layer consists of a $1\times 1$-\myconv layer, that processes the $\mathbf{N}+\mathbf{M}$ to predict the colour distribution $\mathbf{\hat{Z}}\in\mathbb{R}^{H\times W\times Q}$ where $Q$ depicts the dimension of colour distribution over pixels. Finally, the Chroma layer takes $\mathbf{\hat{Z}}$ as input and processes it with a $1\times 1$-\myconv{ } layer upsampling the $\mathbf{\hat{Z}}$ by a factor of 2 and output $(\hat{a},\hat{b})$.

 \subsection{Learning procedure}
\label{cap:learning}
\subsubsection{End-to-end learning}
\label{cap: e2e}
The end-to-end modality of training the model, refers to the standard forward/backward computation of loss gradients passing the input matrix through the entire network from the head till the tail following the Exemplar-based Colourisation~\cite{anwar2020image}.

\subsubsection{Loss Function}
We deal with the multimodality of the colourisation problem applying a Quantisation layer that allows to learn a distribution over quantised pixel colours. We apply quantised colours loss to let the model learn the quantised distribution. The distribution identified by the learnt quantisation is ultimately used to predict the chrominance $(a,b)$ channels for the $\mathbf{L}$ input. This is achieved through the Chroma layer by learning a mapping from the quantised space to the chrominance one by means of the colour error loss.

\paragraph{Quantised colours Loss}
\myalgoname learns a distribution over per-pixel colours. Towards such an objective, inspired by~\cite{zhang2016colorful}, we quantised the $(a,b)$ space into bins with grid size 10 keeping only $Q=313$ values which are in-gamut. These denote the distinct classes a pixel can belong to. Starting from the input channel $\mathbf{L}$, our model learns to generate a distribution over such classes through Quantisation layer and providing $\mathbf{\hat{Z}}$. This is used to compute the quantisation loss
\begin{equation}
    \mathcal{L}_{q} = -\sum_{h,w}v(\mathbf{Z}_{h,w}) \sum_{q}\mathbf{Z}_{h,w,q}log(\mathbf{\hat{Z}}_{h,w,q})
    \label{eq:lq}
\end{equation}
where $\mathbf{Z}_{h,w,q}$ is the ground-truth colour distribution for the $(h,w)$ pixel obtained through a soft-encoding scheme and $v(\cdot)$ re-weights the loss for each pixel based on pixel colour rarity.
We have considered the soft-encoding and the $v(\cdot)$ values introduced by~\cite{zhang2016colorful}.

\paragraph{Colour error loss}
Our final objective is to generate the $(a,b)$ chrominance channels. From $\mathbf{\hat{Z}}$, Chroma layer predicts the $(\hat{a},\hat{b}) \in \mathbb{R}^{H\times W \times 2}$ channels. The colour error loss computes the error obtained by minimising their difference with the real chrominance ones $(a,b)$ as:
\begin{equation}
\mathcal{L}_{c} = ||\hat{a}-a||^2_2 + ||\hat{b}-b||^2_2.
\label{eq:lc}
\end{equation}

\paragraph{Combined Loss}
We optimise our model for $\mathcal{L}_q+\mathcal{L}_c$.
This allows us to generate a plausible colourisation for a same/similar object by exploiting the quantised distribution (learned through $\mathcal{L}_q$) while avoiding the weakness of using $\mathcal{L}_{c}$ alone, which produces de-saturated colours~\cite{zhang2016colorful}.

\begin{figure}[h]
  \centering
  \includegraphics[width=0.5\linewidth]{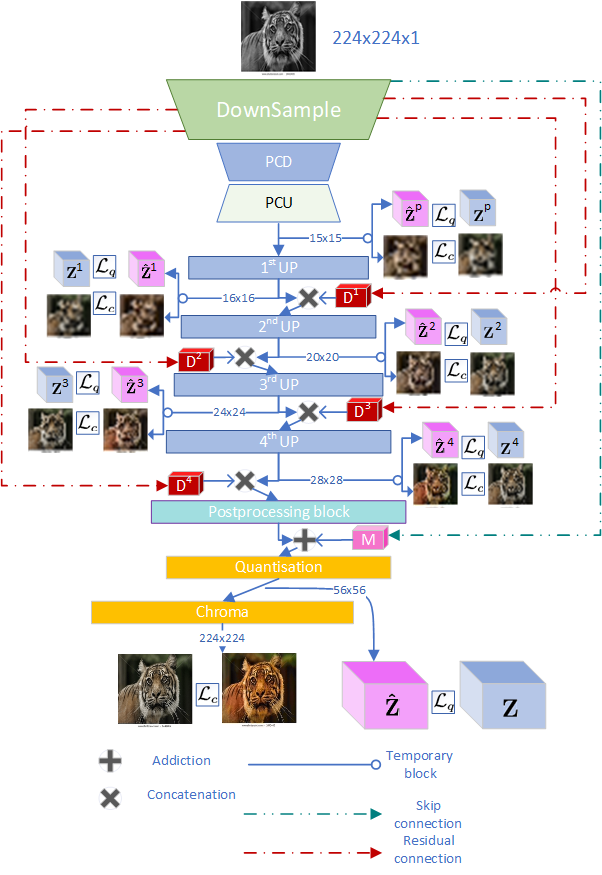}
  \caption{Overall architecture of the proposed progressive learning in downsampling phase. The model learns progresses gradually from first upsampling level to final phase; at each phase the model provides a quantised matrix $q^\alpha$ where $\alpha$ is the upsampling level and a proposed colourisation respect to the resolution of the level $\alpha$.}
  \label{fig:PL loss}
\end{figure}
\subsubsection{Progressive Learning}
A neural network trained with progressive learning develops the ability of incorporating prior knowledge at each layer of the feature hierarchy, reuses old computations and learn new ones. While the training progresses, the network retains a pool of pretrained models throughout training, and learns lateral connections from these to extract useful features. The progressive approach to learning achieves richer compositionality and allows prior knowledge to be integrated at each layer of the feature hierarchy. We apply the idea of progressive learning to \myalgoname to enforce the ability of reconstruction that we are expecting from the decoding phase. We follow the pipeline in Fig.~\ref{fig:PL loss}, where  each $\alpha^{th}$UP layer is added every $\rho$ epochs of learning and the last postprocessing block is trained $\xi$ epochs (the number of epochs used in this work are specified in Section \ref{cap: exp}). At each level of progressive procedure, we train the model to reconstruct the quantisation of colours and the $(a,b)^\alpha$ channels in relation with the resolution of the current level in progression. 
\subsubsection{Progressive Growing Upsampling}
As shown in Fig.~\ref{fig:PL loss}, the number of level in progression is determined by the down-sampling dimension. \myalgoname consists of a preprocessing, four layers of \doubleblockdown{ } with output $\mathbf{D}^\alpha$, a \primarycapsdown, a \primarycapsup, four layers of \doubleblockup{ } with output $\mathbf{Y}^\alpha$, and a final postprocessing block. The input $\mathbf{L}$ is downsampled from size $224\times 224$ to $15\times 15$ and upsampled following the inverted schema. 

In such a structure if trained in end-to-end fashion, the lower phases in upsampling (\primarycapsup, $1^{st}UP,2^{nd}UP$) converge in a quick, stable fashion because there is less colour information. While the resolution increases, \doubleblockup are not able to learn the context of the object due to the limitations of their receptive field and this difficulty affects also the previous layers reducing the benefit of having capsule layers. More over, due to the number of parameters to learn and the amount of details in the ground truth image, the last layer with high resolution is in difficult to learn and extract informative features that affect the backward procedure in training, where the error information about colours at high resolution can be confusing for low resolution. 

In the progressive learning, the lower phases focus on the context of the object, building up a prior knowledge used by the successive layers. This procedure lets the capsules focus on entities features extraction and proposes a solution to the insufficiency of receptive field. In our proposed \myalgoname the architecture of model grows progressively while learning. We add at the current last \doubleblockup{ } a temporary block of two $1\times 1$ convolution layers that provide $\mathbf{\hat{Z}}^\alpha$ and $(\hat{a},\hat{b})^\alpha$ as shown in Fig.~\ref{fig:PL loss}. Every $\rho$ epochs we remove the current temporary block and we add the consecutive \doubleblockup{ } with its relative temporary block. We apply the loss function described in Sec.~\ref{cap:learning}. 
\begin{table*}
  \caption{Detailed metrics: The metrics presented in table are computed for each level of progression in progressive learning (PL). In the last four columns, we present in comparison \myalgoname$_{PL}$ (\myalgoname in progressive learning), \myalgoname in end-to-end. The $^*$ identifies the fine tuning on COCOStuff. We identify with \colorbox{primo}{bright green} the best result in column, \colorbox{secondo}{light green} the second best, and with \colorbox{terzo}{light blue} the third best.}
  \label{tab:stepmetrics}
  \begin{tabular}{l|cccccc|c|c|c}
      \toprule
  ImageNet10k&\primarycapsup{ }&$1^{st}UP$&$2^{nd}UP$&$3^{rd}UP$&$4^{st}UP$&\myalgoname$_{PL}$&\myalgoname$^{*}_{PL}$&\myalgoname&\myalgoname$^{*}$\\
  \midrule
  dim&$15\times 15$&$16\times 16$&$20\times 20$&$24\times 24$&$28\times 28$&$224\times 224$&$224\times 224$&$224\times 224$&$224\times 224$\\
  \midrule
  PSNR$\uparrow$&$31.447$&$31.435$&$31.457$&$31.179$&$31.089$&$31.562$&\cellcolor{primo}$\textbf{32.910}$&\cellcolor{terzo}$31.749$&\cellcolor{secondo}$32.528$\\
  SSIM$\uparrow$&$0.972$&$0.973$&$0.968$&$0.955$&$0.953$&$0.960$&
\cellcolor{primo}$\textbf{0.977}$&\cellcolor{terzo}$0.974$&\cellcolor{secondo}$0.975$\\
  LPIPS$\downarrow$&$0.065$&$0.066$&$0.066$&$0.078$&$0.081$&$0.083$&\cellcolor{primo}$\textbf{0.048}$&\cellcolor{terzo}$0.055$&\cellcolor{secondo}$0.051$\\
  \bottomrule
\end{tabular}
\end{table*}
\begin{figure*}
  \centering
  \includegraphics[width=\textwidth]{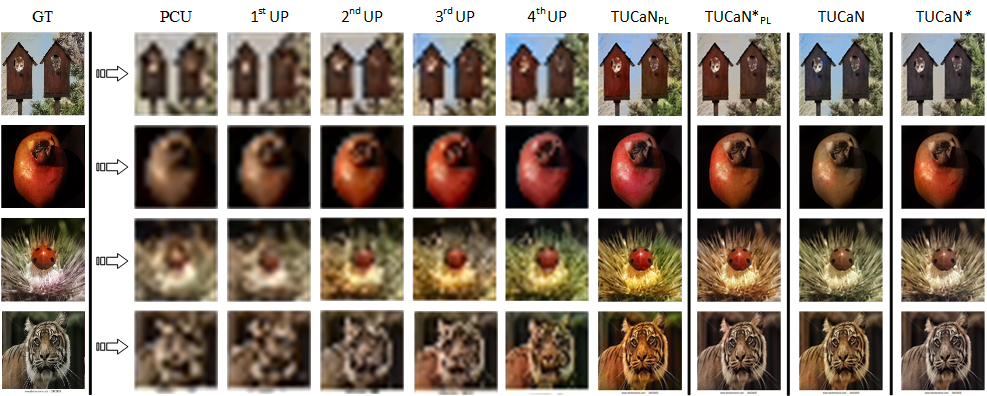}
  \caption{Reconstruction of colours in progressive learning. In the most left column four exemplar images from ImageNet10k dataset; in each of the following column we show the reconstruction result obtained at each level of the progressive learning phase for \myalgoname trained on ImageNet dataset. The \myalgoname$_{PL}$ column is the final colour reconstruction for progressive learning, and the \myalgoname column is the reconstruction for the model trained end-to-end. The "*" identify models fine-tuned on COCOStuff dataset}
  \label{fig:PL samples}
\end{figure*}

\section{Experiments and Results}
\label{cap: exp}
\subsection{Datasets}
To assess the performance of our approach, we have considered three benchmark datasets coming with different features and colourisation challenges.

\textbf{\textit{ImageNet}}~\cite{russakovsky2015imagenet} is widely used as colourisation benchmark.
The 1.3M training images (with no labels) were considered for model training for all the following experiments.
The ImageNet10k~\cite{larsson2016learning} samples have been used for evaluation.

\textbf{\textit{COCOStuff}}~\cite{caesar2018coco} contains a wide variety of natural scenes with multiple objects present in the 118k images. We use the training split for fine tuning procedure and the provided validation split containing 5000 images for evaluation.

\textbf{\textit{Places205}}~\cite{zhou2014learning} is a scene-centric dataset containing samples from 205 different categories. We considered the 20500 validation images for evaluation.

\subsection{Evaluation metrics}
\label{sec:metrics}
To assess the colourisation quality, we followed the experimental protocol proposed in~\cite{Lei_2019_CVPR} and considered the Peak Signal to Noise Ratio (PSNR), the Learned Perceptual Image Patch Similarity (LPIPS)~\cite{zhang2018perceptual} (version 0.1 with VGG backbone), and the Structural Similarity Index Measure (SSIM)~\cite{wang2004image}.
\subsection{Experiments setup\protect\footnote{Code will be publicly released}}
We train the model on the 1.3M ImageNet training samples. The input images are resized to $224\times 224$ and projected into the CIELab colourspace, then the resulting $\mathbf{L}$ channel is used as input of the model. The $(a,b)$ channels are considered to compute the $\mathcal{L}_{q}$ (after soft-encoding) and $\mathcal{L}_{c}$ losses as described in \ref{cap:learning}. 
\paragraph{End-to-end Learning: }We train \myalgoname for 40 epochs, batch size of 32. We apply Adam optimiser with learning rate $2e^{-3}$. Using PyTorch framework, a single epoch of training takes about 4 hours on an NVidia Titan RTX. The model takes 10 minutes to infer the colourisation of the entire COCOStuff dataset.
\paragraph{Progressive Learning:} Following the Progressive learning idea, we train \myalgoname progressively expanding the upsampling phase by a \doubleblockup{ } per time. We use the same dataset parameter used for the end-to-end learning and a similar configuration for the Adam optimiser. We set $\rho=10$ epochs of training in order to give the time to each layer to understand the useful features for the colourisation task. For last expansion we set $\xi=20$ epochs. The training procedure consists of 70 epochs in total. 
\paragraph{Fine tuning}
 We perform the\textit{ fine tuning} of the model on COCOStuff,for 35 epochs with Adam  and learning rate $2e^{-4}$ for \doubleblockdown{ }  and \doubleblockup{ } and learning rate $2e^{-3}$ for \primarycapsdown{ } and \primarycapsup{ }. We experimented robustness in loss descending when applied the differentiation in learning values.

\subsection{Ablation study}
\begin{figure}
  \centering
  \includegraphics[width=0.5\columnwidth]{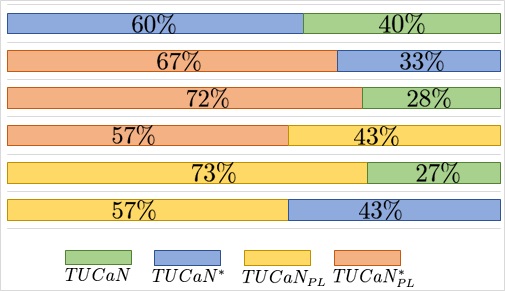}
  \caption{Summary of results obtained user study: we investigate the naturalness of images coloured by \myalgoname trained in end-to-end, or progressive learning $_{PL}$, and w/o fine tuning $"*"$. The percentage reported in table represents the preference of users. }
  \label{fig:userstudy}
\end{figure}

\begin{figure}[h]
  \centering
  \includegraphics[width=0.6\columnwidth]{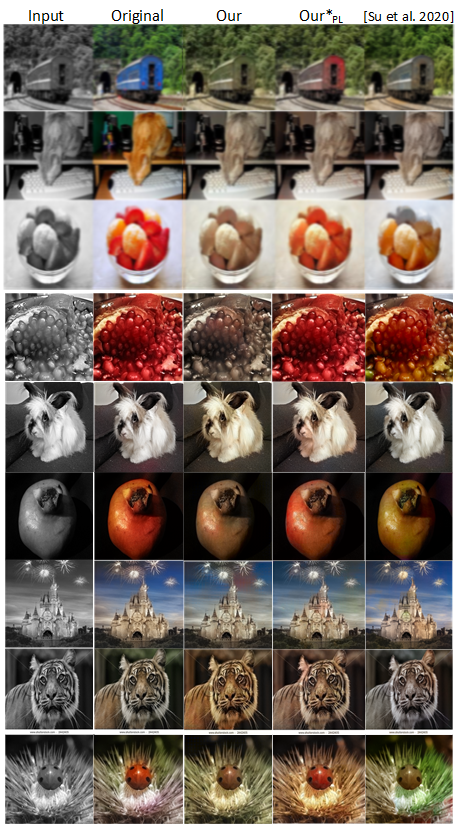}
  \caption{Qualitative comparisons: in figure \myalgoname trained in end-to-end procedure and \myalgoname*$_{PL}$ trained in progressive learning with fine tuning on COCOStuff. The models selected for display are the ones that outperform the SOTA as reported in Tab.~\ref{tab:freq}. The reconstructed colours are compare with original (second column) and \cite{Su2020CVPR} (last column).}
  \label{fig:Compared samples}
\end{figure}

\paragraph{Effectiveness of Progressive Learning:} To investigate the effectiveness of the progressive learning scheme, we conducted ablation study on the images that have entities in foreground that make observable progression in colourisation. In Tab.~\ref{tab:stepmetrics}, we show the quantitative metrics values at each progression step with Image10k dataset. The metrics have a descent ascent behaviour, all of the three metrics provide a better result at level \primarycapsup{ } then they descent in performance while the level increase. This behaviour is ascribable to the increase in complexity of resolution in upsampling, in fact at level \primarycapsup{ } the model predicts colours for an image $15\times 15$ while, growing the upsampling phase, the resolution provided is up to $224\times 224$. This is clearly shown in Fig.~\ref{fig:Compared samples}, where the first and second column in Fig.~\ref{fig:Compared samples} show the colour reconstruction in the \primarycapsup{ } and the first \doubleblockup{ } layers. We note how the colourisation is taking into consideration of the segmented objects through steps in progressive learning. In the last four columns of the table we compare the metrics obtained w/o progressive learning and w/o fine tuning. The metrics demonstrate that the proposed progressive learning for colourisation outperform the application of an end-to-end learning. It is to note the differences in colours reconstruction, the colours predicted by \myalgoname$_{PL}$ are vivid and appealing to humans more than colours predicted by the same model trained in end-to-end fashion. Overall the \myalgoname$*_{PL}$ is qualitatively the best version of \myalgoname.
\paragraph{User study: } We performed a user study to investigate the visual discrepancy between metrics and qualitative colours prediction. The users were asked to choose the more appealing colourisation among the prediction of the four variations of \myalgoname (w/o progressive, w/o fine tuning). We have a statistical sample of 132 users for a total of 2640 answers.  Fig.\ref{fig:userstudy} shows the results obtained. As we observe the models trained in progressive learning predict colourisation that are more appreciated by humans. This is proved by the results shown in Fig.\ref{fig:userstudy}, where the models \myalgoname$_{PL}$ and \myalgoname$*_{PL}$ are preferred respectively by $73\%$ and $72\%$ of the times compared to \myalgoname, and respectively by $57\%$ and $67\%$ of the time if compared to \myalgoname*.

\textit{We want to highlight the discrepancy between the metrics results and the users answers. From Tab.~\ref{tab:stepmetrics}, the \myalgoname trained in end-to-end fashion obtains high performance for all the metrics compared to the \myalgoname$_{PL}$ variation. This result is not confirmed by the user study performed over the same model variations. In fact the \myalgoname$_{PL}$ is preferred to the \myalgoname. This discrepancy point out the reliability of metrics for the colourisation evaluation. In this paper, we report the metrics obtained with our model in parallel to samples of colour reconstructions in order to show up that metrics results and sharpness and appealing colourisation are not always in agreement.}

\begin{table*}
  \caption{Summary of metrics results: we compare metrics obtained with \myalgoname trained in end-to-end and progressive procedures, all the models are trained on ImageNet. We identify with "*", the models fine tuned on COCOStuff. }
  \label{tab:freq}
  \begin{tabular}{l |ccc|ccc|ccc}
    \toprule
    & \multicolumn{3}{c|}{ImageNet ctest10k} & \multicolumn{3}{c|}{COCOStuff validation split} & \multicolumn{3}{c}{Places205 validation split} \\
    \midrule
    & LPIPS$\downarrow$ & PSNR$\uparrow$ & SSIM$\uparrow$ & LPIPS$\downarrow$ & PSNR$\uparrow$ & SSIM$\uparrow$  &  LPIPS$\downarrow$ & PSNR$\uparrow$ & SSIM$\uparrow$  \\
    \midrule
    \myalgoname & 
    \cellcolor{primo}0.055& \cellcolor{primo}31.749& \cellcolor{primo}0.974  & \cellcolor{secondo}0.137& \cellcolor{primo}30.366& \cellcolor{terzo}0.918  &
     \cellcolor{secondo}0.136& \cellcolor{primo}30.450&0.923\\
    \myalgoname$_{PL}$& \cellcolor{secondo}0.083& \cellcolor{secondo}31.562&\cellcolor{secondo}0.960&
    0.155&\cellcolor{secondo}30.273&0.905&
    0.152&\cellcolor{secondo}30.323&0.909\\
    \midrule
    Larsson~\cite{larsson2016learning} & 0.188&24.93&0.927&0.183&25.06&0.930&0.161&25.72&\cellcolor{secondo}0.951\\
    Iizuka~\cite{iizuka2016let}&0.200&23.63&0.917&0.185&23.86&0.917&\cellcolor{terzo}0.146&25.58&0.950\\
    Zhang~\cite{zhang2017real}&0.145  &26.166&0.932&\cellcolor{terzo}0.138&26.823&\cellcolor{secondo}0.937&0.149&25.823&0.948\\
    Su~\cite{Su2020CVPR} &\cellcolor{terzo}0.134&\cellcolor{terzo}26.98&\cellcolor{terzo}0.933& \cellcolor{primo}0.125& \cellcolor{terzo}27.77& \cellcolor{primo}0.940  &\cellcolor{primo}0.130&\cellcolor{terzo}27.16&\cellcolor{primo}0.954\\
    \midrule
    \midrule
    \myalgoname* & \cellcolor{secondo}0.051&\cellcolor{secondo}32.528&\cellcolor{secondo}0.975&\cellcolor{terzo}0.128&\cellcolor{secondo}30.511&0.920&\cellcolor{terzo}0.127&\cellcolor{secondo}30.661&0.926\\
    \myalgoname*$_{PL}$ & \cellcolor{primo}0.048&\cellcolor{primo}32.910&\cellcolor{primo}0.977&\cellcolor{secondo}0.126&\cellcolor{primo}30.713&\cellcolor{terzo}0.921& 0.126&\cellcolor{primo}30.711&\cellcolor{terzo}0.926\\
    \midrule
    Zhang *~\cite{zhang2017real}&0.140&26.482&0.932&0.128&27.251&\cellcolor{secondo}0.938&\cellcolor{secondo}0.153&25.720&\cellcolor{secondo}0.947\\
    Su *~\cite{Su2020CVPR}&\cellcolor{terzo}0.125&\cellcolor{terzo}27.562&\cellcolor{terzo}0.937&\cellcolor{primo}0.110&\cellcolor{terzo}28.592&\cellcolor{primo}0.944&\cellcolor{primo}0.120&\cellcolor{terzo}27.800&\cellcolor{primo}0.957\\
    
  \bottomrule
\end{tabular}
\end{table*}

\subsection{Comparison with the state of the art}
In this section, we compare \myalgoname results with results at the SOTA. We present a quantitative analysis that takes into consideration results from metrics and a qualitative analysis that present visually colours reconstruction. For the quantitative analysis, we take into consideration latest models in colourisation~\cite{larsson2016learning,iizuka2016let,zhang2017real,Su2020CVPR}. We analyse separately metrics from \cite{ozbulak2019image}; this division is due to the application of different test sets. For the qualitative, analysis we present colours reconstructed with the closest models for colourisation \cite{Su2020CVPR,ozbulak2019image}.

\subsubsection{Quantitative comparisons}
In Tab.~\ref{tab:freq}, the results with \myalgoname refer to end-to-end learning, and \myalgoname$_{PL}$ refer to progressive learning, finally the models with "*" are fine tuned.
We observe that \myalgoname in end-to-end outperforms the SOTA when tested on ImageNet10K. In fact compared to the SOTA, the LPIPS value obtained is reduced to one-third, SSIM is improved by 0.041 and PSNR by 4.769. We obtain good results also with \myalgoname$_{PL}$, the LPIPS is reduced by 0.035, the PSNR increased by 4.582 , and SSIM increased by 0.027 compared to \cite{Su2020CVPR}. The results with COCOStuff and Places205 dataset are promising placing \myalgoname right below the best at the SOTA for LIPS and SSIM and obtaining the best value for PSNR. These results prove \myalgoname as competitive with \cite{Su2020CVPR}. We believe that this change in behaviour is mainly due to the nature of the definition of images in the datasets, while ImageNet10k consists of foreground subjects, COCOStuff and Places205 consists of landscapes and scenes.
Performing the fine tuning, we observe a similar behaviour as before. The performance obtained by \myalgoname*$_{PL}$ for ImageNet10k outperforms the proposed SOTA in all the metrics obtaining outstanding LPIPS and SSIM values. We compare the metrics obtained by \myalgoname*$_{PL}$ and \myalgoname -the two best version of \myalgoname according to Tab.\ref{tab:freq} - with \cite{ozbulak2019image,Su2020CVPR} with the DIV2K dataset~\cite{Agustsson_2017_CVPR_Workshops}. Following the metrics presented in \cite{ozbulak2019image}, we compare the models over the PSNR and SSIM metrics.
\begin{table}
  \caption{Metrics results on DIV2K dataset \cite{Agustsson_2017_CVPR_Workshops}: we compare \myalgoname, and \myalgoname*$_{PL}$ with the two latest and closest works at the SOTA - \cite{ozbulak2019image,Su2020CVPR}.}
  \label{tab:div2kmetrics}
  \begin{tabular}{l |cc}
  &PSNR$\uparrow$&SSIM$\uparrow$\\
  \midrule
    \myalgoname&\cellcolor{secondo}30.39&\cellcolor{secondo}0.907\\
    \myalgoname*$_{PL}$&\cellcolor{primo}30.40&\cellcolor{primo}0.910\\	
    \cite{Su2020CVPR}&\cellcolor{terzo}30.042&\cellcolor{terzo}0.886\\	
    \cite{ozbulak2019image}&21.08&0.85	
  
\end{tabular}
\end{table}
Results presented in Tab.\ref{tab:div2kmetrics} show that \myalgoname in both versions outperforms the results obtained with considered models. These prove the improvement achieved with \myalgoname compared to the SOTA of capsules \cite{ozbulak2019image} in colourisation and confirm the competitive behaviour compared to \cite{Su2020CVPR}
\subsubsection{Qualitative comparisons}
Fig.~\ref{fig:Compared samples} shows some samples of reconstructed images from COCOStuff -first three rows-  and ImageNet10k -last six rows- with \myalgoname, \myalgoname*$_{PL}$ and the colourisation obtained with \cite{Su2020CVPR}. We observe that compared to \cite{Su2020CVPR}, \myalgoname*$_{PL}$ provides more vivid colourisation, and the colours appear natural and sharp. This qualities are more evident for the ImageNet10k images following the metrics in Tab.~\ref{tab:freq}. In the second and third columns referred to the two variation of \myalgoname, we observe that the more appealing colourisations are obtained with \myalgoname*$_{PL}$. By observation, we believe that \myalgoname provides a colourisation that is well defined for at global level for each image and for entities present in them.  
\begin{figure}[h]
  \centering
  \includegraphics[width=0.6\columnwidth]{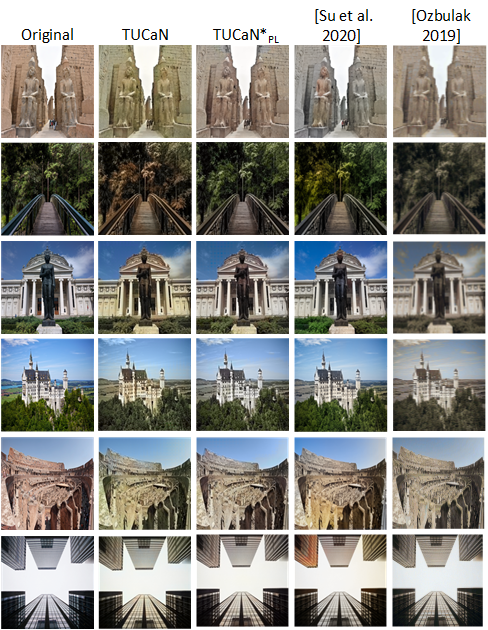}
  \caption{Qualitative comparison with works closest to \myalgoname on DIV2K dataset \cite{Agustsson_2017_CVPR_Workshops}. Here we take into consideration \myalgoname,  and \myalgoname*$_{PL}$, that demonstrate in Tab.\ref{tab:stepmetrics} the two best variations of \myalgoname. We compare \myalgoname*$_{PL}$ with \cite{ozbulak2019image} and \cite{Su2020CVPR} considered the closest for intent and structures to our proposed work.}
  \label{fig:cap_ia}
\end{figure}
 In Fig.~\ref{fig:cap_ia}, we show the predicted colourisation for \myalgoname, and \myalgoname*$_{PL}$ with \cite{ozbulak2019image,Su2020CVPR}. We observe that compared to \cite{ozbulak2019image}, \myalgoname provides vibrant and sharp colourisation without grey areas. With \cite{Su2020CVPR} the colours are sharp and compared to them, \myalgoname*$_{PL}$ achieves homogeneous colourisation and  consistency on colours. This is mainly observable on the first and last rows where \cite{Su2020CVPR} provide a grey colourisation and stains of colours while \myalgoname achieves a consistent colourisation.

\section{Conclusions}
In this paper, we present the \myalgoname architecture that combining convolutional and capsules layers is designed to solve the ill-posed image colourisation problem. The proposed method is fully automatic, end-to-end, and it exploits the application of convolutional layers for image-level features extraction and entity-level feature extraction by capsules layers for the colourisation task. We propose \myalgoname trained with the progressive learning scheme that progressively trains the upsampling from low-resolution to high-resolution, where the lower phases weakly colourise the entities and higher phases refine the sharpness of colours. Compared with existing state-of-the-art methods, we encourage the capture of colours by the growth of architecture obtained with the manipulation of the learning scheme.

Experiments show that the \myalgoname outperforms results on ImageNet10k compared to the SOTA and obtains outperformed and competitive results on COCOStuff and Places205 datasets. We compare \myalgoname with previous application of capsule for colourisation and the result of the comparison demonstrate that the proposes method outperforms SOTA for capsules. We demonstrate that \myalgoname trained with progressive learning is a valid model for colourisation of greyscale images. We also prove that the application of progressive learning for colourisation achieve better qualitative and quantitative performance then the end-to-end learning procedure.

\bibliographystyle{unsrt}  
\bibliography{template}  

\begin{thebibliography}{10}

\bibitem{cheng2015deep}
Zezhou Cheng, Qingxiong Yang, and Bin Sheng.
\newblock Deep colorization.
\newblock In {\em Proceedings of the IEEE International Conference on Computer
  Vision}, pages 415--423, 2015.

\bibitem{Su2020CVPR}
Jheng-Wei Su, Hung-Kuo Chu, and Jia-Bin Huang.
\newblock Instance-aware image colorization.
\newblock In {\em Proceedings of the IEEE/CVF Conference on Computer Vision and
  Pattern Recognition (CVPR)}, June 2020.

\bibitem{iizuka2016let}
Satoshi Iizuka, Edgar Simo-Serra, and Hiroshi Ishikawa.
\newblock Let there be color! joint end-to-end learning of global and local
  image priors for automatic image colorization with simultaneous
  classification.
\newblock {\em ACM Transactions on Graphics (ToG)}, 35(4):1--11, 2016.

\bibitem{zhang2016colorful}
Richard Zhang, Phillip Isola, and Alexei~A Efros.
\newblock Colorful image colorization.
\newblock In {\em European conference on computer vision}, pages 649--666.
  Springer, 2016.

\bibitem{larsson2016learning}
Gustav Larsson, Michael Maire, and Gregory Shakhnarovich.
\newblock Learning representations for automatic colorization.
\newblock In {\em European conference on computer vision}, pages 577--593.
  Springer, 2016.

\bibitem{deshpande2017learning}
Aditya Deshpande, Jiajun Lu, Mao-Chuang Yeh, Min Jin~Chong, and David Forsyth.
\newblock Learning diverse image colorization.
\newblock In {\em Proceedings of the IEEE Conference on Computer Vision and
  Pattern Recognition}, pages 6837--6845, 2017.

\bibitem{guadarrama2017pixcolor}
Sergio Guadarrama, Ryan Dahl, David Bieber, Mohammad Norouzi, Jonathon Shlens,
  and Kevin Murphy.
\newblock Pixcolor: Pixel recursive colorization.
\newblock {\em arXiv preprint arXiv:1705.07208}, 2017.

\bibitem{isola2017image}
Phillip Isola, Jun-Yan Zhu, Tinghui Zhou, and Alexei~A Efros.
\newblock Image-to-image translation with conditional adversarial networks.
\newblock In {\em Proceedings of the IEEE conference on computer vision and
  pattern recognition}, pages 1125--1134, 2017.

\bibitem{he2018deep}
Mingming He, Dongdong Chen, Jing Liao, Pedro~V Sander, and Lu~Yuan.
\newblock Deep exemplar-based colorization.
\newblock {\em ACM Transactions on Graphics (TOG)}, 37(4):1--16, 2018.

\bibitem{ozbulak2019image}
Gokhan Ozbulak.
\newblock Image colorization by capsule networks.
\newblock In {\em Proceedings of the IEEE Conference on Computer Vision and
  Pattern Recognition Workshops}, pages 0--0, 2019.

\bibitem{mouzon2019joint}
Thomas Mouzon, Fabien Pierre, and Marie-Odile Berger.
\newblock Joint cnn and variational model for fully-automatic image
  colorization.
\newblock In {\em International Conference on Scale Space and Variational
  Methods in Computer Vision}, pages 535--546. Springer, 2019.

\bibitem{sabour2017dynamic}
Sara Sabour, Nicholas Frosst, and Geoffrey~E Hinton.
\newblock Dynamic routing between capsules.
\newblock In {\em Advances in neural information processing systems}, pages
  3856--3866, 2017.

\bibitem{karras2017progressive}
Tero Karras, Timo Aila, Samuli Laine, and Jaakko Lehtinen.
\newblock Progressive growing of gans for improved quality, stability, and
  variation.
\newblock {\em arXiv preprint arXiv:1710.10196}, 2017.

\bibitem{russakovsky2015imagenet}
Olga Russakovsky, Jia Deng, Hao Su, Jonathan Krause, Sanjeev Satheesh, Sean Ma,
  Zhiheng Huang, Andrej Karpathy, Aditya Khosla, Michael Bernstein, et~al.
\newblock Imagenet large scale visual recognition challenge.
\newblock {\em International journal of computer vision}, 115(3):211--252,
  2015.

\bibitem{caesar2018coco}
Holger Caesar, Jasper Uijlings, and Vittorio Ferrari.
\newblock Coco-stuff: Thing and stuff classes in context.
\newblock In {\em Proceedings of the IEEE Conference on Computer Vision and
  Pattern Recognition}, pages 1209--1218, 2018.

\bibitem{zhou2014learning}
Bolei Zhou, Agata Lapedriza, Jianxiong Xiao, Antonio Torralba, and Aude Oliva.
\newblock Learning deep features for scene recognition using places database.
\newblock In {\em Advances in neural information processing systems}, pages
  487--495, 2014.

\bibitem{nazeri2018image}
Kamyar Nazeri, Eric Ng, and Mehran Ebrahimi.
\newblock Image colorization using generative adversarial networks.
\newblock In {\em International conference on articulated motion and deformable
  objects}, pages 85--94. Springer, 2018.

\bibitem{vitoria2020chromagan}
Patricia Vitoria, Lara Raad, and Coloma Ballester.
\newblock Chromagan: Adversarial picture colorization with semantic class
  distribution.
\newblock In {\em The IEEE Winter Conference on Applications of Computer
  Vision}, pages 2445--2454, 2020.

\bibitem{zhao2018pixel}
Jiaojiao Zhao, Li~Liu, Cees~GM Snoek, Jungong Han, and Ling Shao.
\newblock Pixel-level semantics guided image colorization.
\newblock {\em arXiv preprint arXiv:1808.01597}, 2018.

\bibitem{zhang2017real}
Richard Zhang, Jun-Yan Zhu, Phillip Isola, Xinyang Geng, Angela~S Lin, Tianhe
  Yu, and Alexei~A Efros.
\newblock Real-time user-guided image colorization with learned deep priors.
\newblock {\em arXiv preprint arXiv:1705.02999}, 2017.

\bibitem{natsumi2019}
Natsumi Kato, Hiroyuki Osone, Kotaro Oomori, Chun~Wei Ooi, and Yoichi Ochiai.
\newblock Gans-based clothes design: Pattern maker is all you need to design
  clothing.
\newblock In {\em Proceedings of the 10th Augmented Human International
  Conference 2019}, 2019.

\bibitem{kang2020pl}
Joonki~Paik Dong-Goo~Kang, Sangwoo~Park.
\newblock Coarse to fine: Progressive and multi-task learning for salient
  object detection.
\newblock In {\em 2020 - International conference of pattern recognition},
  2020.

\bibitem{hadji2018we}
Isma Hadji and Richard~P Wildes.
\newblock What do we understand about convolutional networks?
\newblock {\em arXiv preprint arXiv:1803.08834}, 2018.

\bibitem{anwar2020image}
Saeed Anwar, Muhammad Tahir, Chongyi Li, Ajmal Mian, Fahad~Shahbaz Khan, and
  Abdul~Wahab Muzaffar.
\newblock Image colorization: A survey and dataset.
\newblock {\em arXiv preprint arXiv:2008.10774}, 2020.

\bibitem{Lei_2019_CVPR}
Chenyang Lei and Qifeng Chen.
\newblock Fully automatic video colorization with self-regularization and
  diversity.
\newblock In {\em The IEEE Conference on Computer Vision and Pattern
  Recognition (CVPR)}, June 2019.

\bibitem{zhang2018perceptual}
Richard Zhang, Phillip Isola, Alexei~A Efros, Eli Shechtman, and Oliver Wang.
\newblock The unreasonable effectiveness of deep features as a perceptual
  metric.
\newblock In {\em CVPR}, 2018.

\bibitem{wang2004image}
Zhou Wang, Alan~C Bovik, Hamid~R Sheikh, and Eero~P Simoncelli.
\newblock Image quality assessment: from error visibility to structural
  similarity.
\newblock {\em IEEE transactions on image processing}, 13(4):600--612, 2004.

\bibitem{Agustsson_2017_CVPR_Workshops}
Eirikur Agustsson and Radu Timofte.
\newblock Ntire 2017 challenge on single image super-resolution: Dataset and
  study.
\newblock In {\em The IEEE Conference on Computer Vision and Pattern
  Recognition (CVPR) Workshops}, July 2017.

\end{thebibliography}






\end{document}